\documentclass{article}

\usepackage{arxiv}

\usepackage[utf8]{inputenc} 
\usepackage[T1]{fontenc}    
\usepackage{hyperref}       
\usepackage{url}            
\usepackage{booktabs}       
\usepackage{amsfonts}       
\usepackage{nicefrac}       
\usepackage{microtype}      
\usepackage{graphicx}
\usepackage[numbers]{natbib}
\usepackage{xcolor}
\usepackage{tikz}
\usepackage{amsmath}
\usepackage{wrapfig}  
\usepackage{float}
\usepackage[normalem]{ulem}

\newcommand{\minihead}[1]{\textsc{\textbf{#1}}\hspace{0.5em}}

\title{MUMU: Bootstrapping Multimodal Image Generation from Text-to-Image Data}
\author{William Berman \\
Researcher in Residence \\
Sutter Hill Ventures \\
\texttt{WLBberman@gmail.com}
\And
Alex Peysakhovich\thanks{Corresponding author \\ \\
All face images in this work are used with permission. All non-generated images images in this work are owned by their respective copyright holders. All images are used only for research purposes.} \\
Researcher in Residence \\
Sutter Hill Ventures \\
\texttt{alex.peys@gmail.com}
}

\renewcommand{\shorttitle}{MUMU}

\begin{document}
\maketitle

\begin{abstract}

We train a model to generate images from multimodal prompts of interleaved text and images such as “a $\boldsymbol{<}$picture of a man$\boldsymbol{>}$ man and his $\boldsymbol{<}$picture of a dog$\boldsymbol{>}$ dog in an $\boldsymbol{<}$picture of a cartoon$\boldsymbol{>}$ animated style.” We bootstrap a multimodal dataset by extracting semantically meaningful image crops corresponding to words in the image captions of synthetically generated and publicly available text-image data. Our model, MUMU, is composed of a vision-language model encoder with a diffusion decoder and is trained on a single 8xH100 GPU node. Despite being only trained on crops from the same image, MUMU learns to compose inputs from different images into a coherent output. For example, an input of a realistic person and a cartoon will output the same person in the cartoon style, and an input of a standing subject and a scooter will output the subject riding the scooter. As a result, our model generalizes to tasks such as style transfer and character consistency. Our results show the promise of using multimodal models as general purpose controllers for image generation.
\end{abstract}

\begin{figure}[ht!]
    \centering
    \includegraphics[scale=.28]{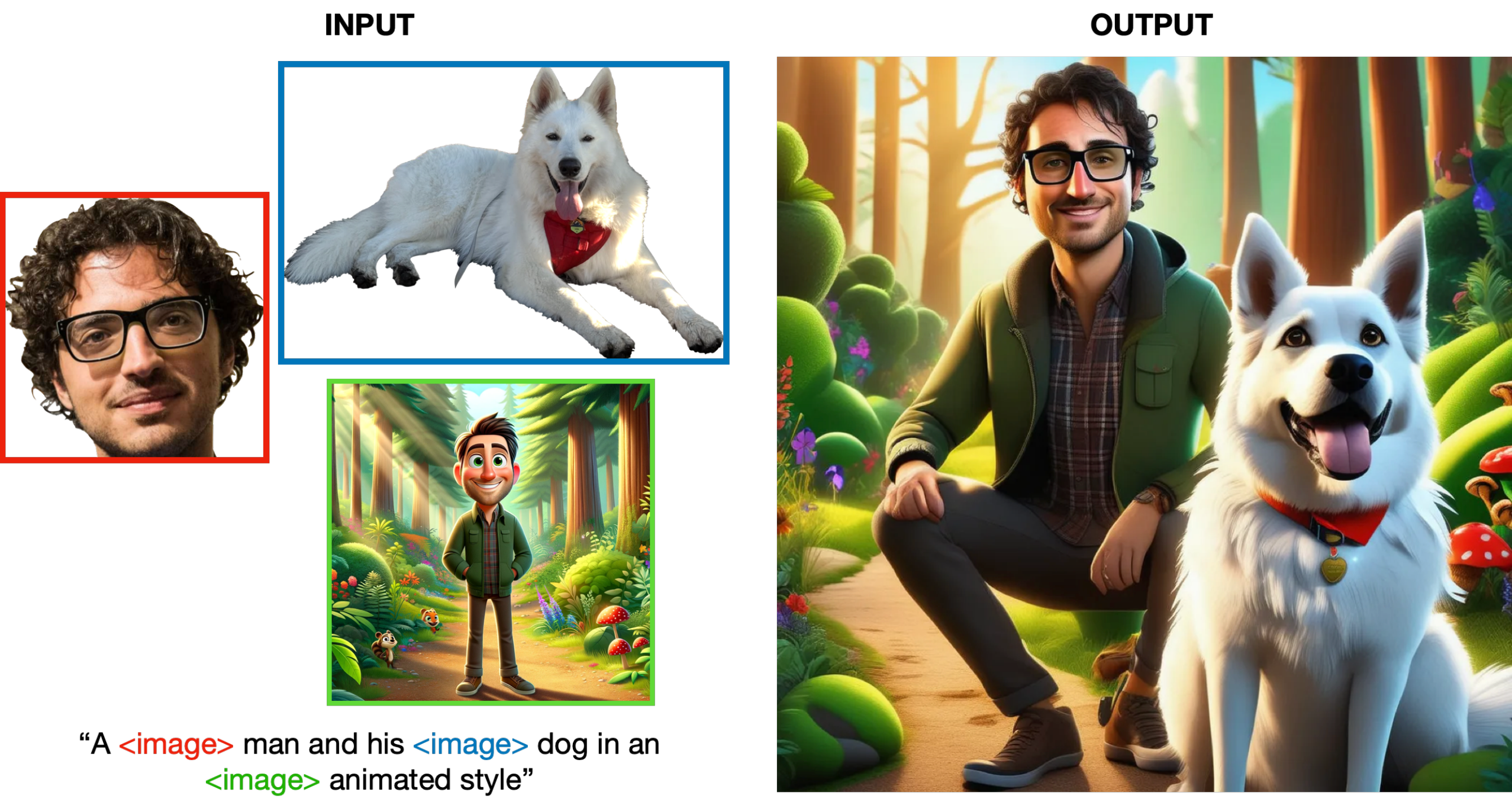}
    \caption{An example of a multimodal prompt, and a resulting generation from our MUMU-Idefics2-SDXL model. The model inputs multimodal conditioning and outputs images.}
    \label{fig:first_example}
\end{figure}


\section{Introduction}
Text-to-image generative AI produces detailed images from simple text prompts \citep{rombach2022high,yu2022scaling,betker2023improving,saharia2022photorealistic,dai2023emu,podell2023sdxl,esser2024scaling}. However, text is not always sufficient to describe user intent. One potential improvement is multimodal prompting which allows users to specify both text and reference images. We bootstrap a multimodal prompt dataset from existing text-image data, and we train our image generation model with multimodal understanding, MUMU, by replacing the text encoder, CLIP \citep{radford2021learning}, of a diffusion model \citep{sohldickstein2015deep,ho2020denoising,song2020denoising,song2020generative, song2021scorebased, karras2022elucidating}, SDXL \citep{podell2023sdxl}, with a vision-language model \citep{alayrac2022flamingo,chen2022visualgpt,liu2024visual,liu2023improved,laurenccon2024matters,laurenccon2024obelics,bordes2024introduction}, Idefics2 \citep{laurenccon2024matters}.

\begin{wrapfigure}{r}{0.5\textwidth}  
  \centering
  \includegraphics[scale=.33]{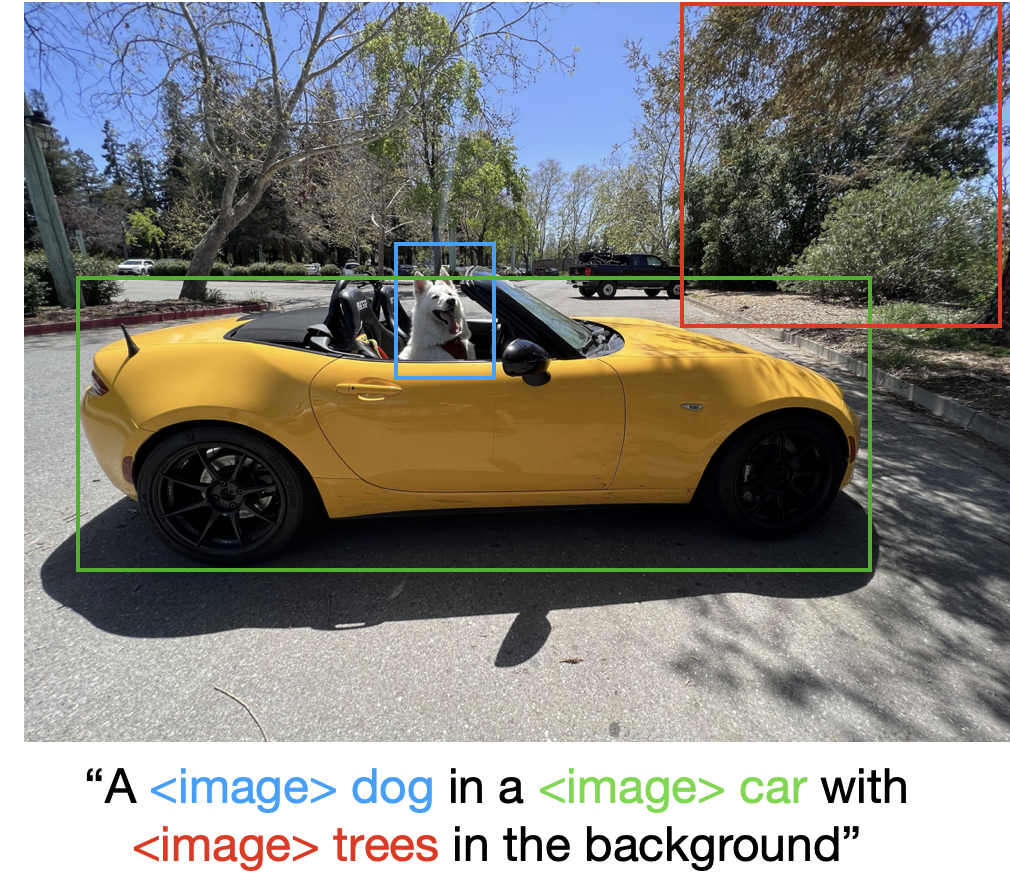}
  \caption{
  A stylized example (not from our dataset) of the multimodal caption for a text-image pair. The object detection bounding boxes are cropped and inserted into the multimodal prompt before their corresponding words.}
  \label{fig:mumudata}
\end{wrapfigure}

We construct a multimodal training set bootstrapped from text-image data. We use open vocabulary object detection to extract image crops corresponding to words in the image captions \citep{liu2023grounding}. The image crops are then placed before their corresponding words in the text prompt, see Figure \ref{fig:mumudata} for an example. Our data is mostly synthetically generated from SDXL with some added high quality publicly available data.

SDXL conditions on text via cross-attention \citep{vaswani2017attention} on CLIP hidden states. We replace the CLIP hidden states with those of a minorly modified Idefics2. Idefics2 is composed of a vision transformer \citep{dosovitskiy2020image} for embedding image inputs, a perceiver transformer for pooling image embeddings to a fixed sequence length, and a large vision-language model transformer. For MUMU, we remove Idefics2's perceiver transformer to use a larger number of tokens per image. We find that removing the perceiver and using more tokens improves image quality with image quality saturating at approximately $1,000$ tokens per image. We also add a small non-causal ``adapter'' transformer on top of Idefics2's hidden states \citep{hu2024ella}. Figure \ref{fig:arch} shows the full architecture. MUMU is further trained from base SDXL and Idefics2 on a single 8xH100 GPU node for approximately 300,000 steps or 6 days.  

MUMU can directly place conditioning images into the generated image. Additionally, despite being only trained on crops from the same input image, MUMU can also harmonize conditioning images from different inputs into a coherent output. E.g. an input of a realistic person and a cartoon will output the same person in the cartoon style, see Figure \ref{fig:first_example}. As a result, the model generalizes to tasks such as style transfer and character consistency. Additionally, the model can be merged into style-based fine-tunes of SDXL to generate conditioning objects in the fine-tune's style.

MUMU struggles with consistency of small details (e.g. fine details of faces or clothing) and has slightly worse pure text adherence than base SDXL. Additionally, canonical Stable Diffusion artifacts carry over to MUMU e.g. `bleeding' of conditioning across multiple objects. Some of these issues can likely be solved by scaling or by more careful dataset construction while other issues may require specialized design decisions.

In this report, we discuss MUMU's dataset construction, strengths, and weaknesses. We also discuss how MUMU-like architectures fit into the general problem of controllable image generation.

\section{Related Work and Background}
\subsection{Diffusion Models}

Diffusion models generate images by iteratively applying a denoising procedure. There are many ways to construct diffusion models, and choices such as how noise is added, how noise is removed, prediction targets, and model architecture are all well studied \citep{sohldickstein2015deep,ho2020denoising,song2020denoising,song2020generative, song2021scorebased, karras2022elucidating, lu2022dpmsolver,rombach2022high,podell2023sdxl,saharia2022photorealistic,chen2023pixartalpha}.

For this work, we use SDXL for which we give a quick summary and point the interested reader to the original paper for more details \citep{podell2023sdxl}. During training, image caption pairs, $(p^i, c^i)$ are respectively encoded with a variational auto-encoder \citep{rombach2022high,esser2021taming,kingma2013auto,razavi2019generating}, $x^i_0 = V(p^i)$, and a text encoder. A noise schedule is parameterized by noise levels, $t$, with variances, $\beta_t$, such that a noised $x^i_t$ is sampled as $x^i_t \sim N(x^i_0 \sqrt{1-\beta_t}, \beta_t \mathbf{I})$ i.e. $x^i_t = x_0 \sqrt{1-\beta_t} + \epsilon^i \sqrt{\beta_t}$, $\epsilon^i \sim N(\mathbf{0}, \mathbf{I})$. The SDXL UNet, $m$, predicts $\hat{\epsilon^i} = m(x^i_t, c^i, t)$ and is trained with MSE over batches i.e. $L = \frac{1}{n} \sum_i (\hat{\epsilon^i} - \epsilon^i)^2$.

For inference on a new caption $c^j$, an initial latent, $x^j_T \sim N(\mathbf{0}, \mathbf{I})$, is incrementally denoised over a subset of all timesteps to predict an unnoised latent, $x^j_0$, which is decoded to pixel space by the VAE. For this paper, we use 50 step DDIM \cite{song2020denoising} and classifier free guidance \citep{ho2022classifier,dhariwal2021diffusion} for all predictions.

SDXL conditions on the text, $c$, via cross attention \cite{vaswani2017attention} on CLIP \cite{radford2021learning} embeddings. There is evidence in favor of replacing the CLIP checkpoints used by SD1.5 and SDXL with larger encoders \citep{saharia2022photorealistic,hu2024ella,esser2024scaling,chen2024pixartsigma}. Our main contribution is replacing CLIP with a VLM and allowing $c$ to be multimodal.

\subsection{Vision Language Models (VLMs)}
Language models operate on hidden states from tokenized text while vision transformers operate on hidden states from directly projected \cite{dosovitskiy2020image} or tokenized \cite{oord2018neural} image patches. VLMs operate on the interleaved hidden states of both text and images \citep{alayrac2022flamingo,chen2022visualgpt,liu2024visual,liu2023improved,laurenccon2024obelics,laurenccon2024matters}. Much VLM research has focused on adding image inputs to the same next-token-prediction decoder. Results show that even relatively small language models augmented with image inputs and fine tuned on a relatively small dataset can become capable visual reasoners in tasks like visual question answering \cite{liu2023improved}.

MUMU's encoder uses a standard vision-language architecture, see Figure \ref{fig:arch}. Our main results complement the VLM literature by showing that not only can existing \textit{encoders} be easily bolted onto a pre-trained language module but existing \textit{decoders} can be as well.

\subsection{Natively Multimodal Models}
Open VLMs usually hybridize two pre-trained backbones. Existing work has demonstrated the image generation capabilities of natively trained multimodal models \cite{team2023gemini,hu2024instruct,openai2024gpt4o}. However, these models are closed weights and cannot be directly built upon. Our work shows that a multimodal image generation model can be constructed from existing open weights models allowing for community research on this topic.

\subsection{Diffusion Model Image Conditioning}
We are not the first to point out that text is an insufficient vehicle for user intent in image generation. ControlNet \citep{zhang2023adding}, T2I-Adapter \citep{mou2023t2iadapter}, and IP-Adapter \citep{ye2023ip} are non-mutually exclusive \citep{wang2024instantid} methods of adding auxiliary image conditioning to pre-trained diffusion models. For all three methods, the output of a conditioning specific encoder is injected through either residual connections (ControlNet, T2I-Adapter) or newly added cross attention modules (IP-Adapter). Depth map, canny edge, and pose are examples of common ControlNet and T2I-Adapter conditionings, and reference image, style, character, and clothing are common IP-Adapter conditionings \citep{seyfioglu2024diffuse,huang2024consistentid,wang2024instantstyle}.

Training separate encoders per auxiliary conditioning has been useful for experimenting with and cheaply using new conditionings on top of a common base model. However, auxiliary encoders trained separately on top of a frozen base model can require inference time parameter tuning \citep{zhao2024uni}, and many added encoders may compose poorly.

Prior work \citep{pan2023kosmos} has considered combining relatively small multimodal encoders with SD1.5 using a similar training idea to the one presented here (object detection and interleaving captions) along with an alignment and instruction training phase. We add to this work by showing that just the interleaved object pre-training phase along with scaling the encoder and decoder is enough to achieve high quality multimodal control. In addition, we argue that training the entire model end-to-end rather than freezing decoders yields better outcomes.

\subsection{Style Transfer}
Image style transfer has been studied across machine learning \cite{jing2019neural,deng2020arbitrary, an2021artflow,wang2024instantstyle,sohn2023styledrop}. Inversion \cite{deng2020arbitrary, an2021artflow}, embedding injection, and block specific embedding injection \citep{wang2024instantstyle,sohn2023styledrop,rout2024rbmodulation} have all been used for style transfer in diffusion models. In comparison, multimodal encoders can be directly prompted with style reference images and do not require specialized architectures.

\section{MUMU Architecture}

We start from SDXL's pre-trained convolutional UNet with transformer blocks cross attending to CLIP hidden states. We then ablate SDXL's auxiliary CLIP text encoder and replace SDXL's primary CLIP text encoder with the hidden states of the VLM Idefics2 \citep{laurenccon2024matters}.

\begin{figure}[h]
    \centering
    \includegraphics[scale=.3]{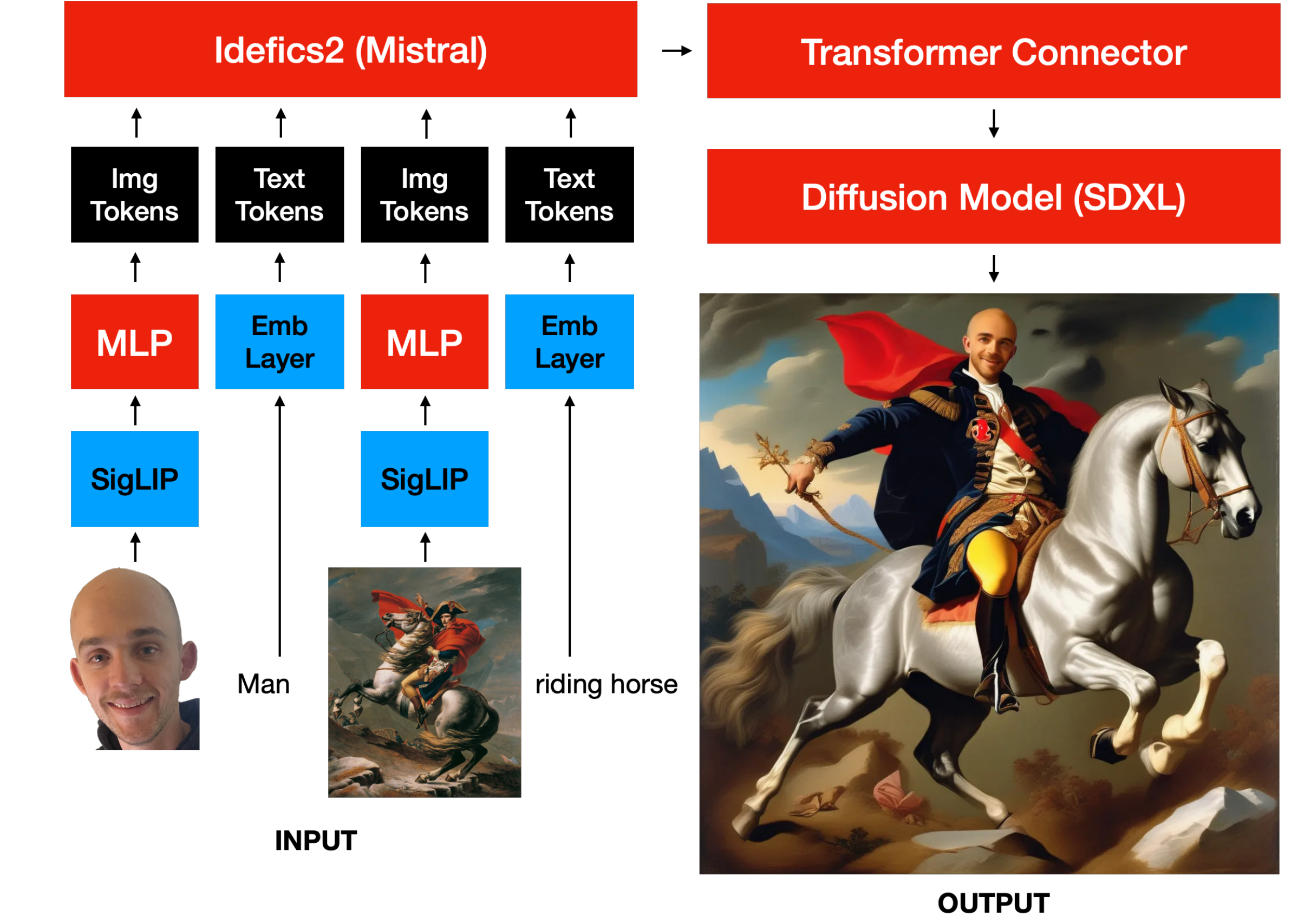}
    \caption{MUMU-Idefics2-SDXL architecture. Red indicates modules which are trained, blue indicates frozen, black indicates embedding. Output is actual output from MUMU-Idefics2-SDXL to the given prompt.}
    \label{fig:arch}
\end{figure}

Idefics2 is composed of a vision transformer \citep{dosovitskiy2020image} initialized from SigLIP \citep{zhai2023sigmoid} for embedding image inputs, a perceiver transformer for pooling image embeddings to a fixed sequence length \citep{jaegle2021perceiver}, and a large vision-language model transformer initialized from Mistral 7b \cite{jiang2023mistral}. We ablate the perceiver transformer because we find its small number of image tokens hurt detail preservation. The Idefics2 hidden states are passed to a 16 layer non-causal transformer and then into SDXL cross-attention. Figure \ref{fig:arch} shows the architecture. 

\section{Constructing MUMU Captioned Datasets}

\minihead{object extraction} We bootstrapped our multimodal prompt dataset from text-image data by using GroundingDINO’s open vocab object detection \citep{liu2023grounding} to extract image crops corresponding to words in image captions. We filtered out crops with total area less than $150 \times 150$ pixels or larger than $75\%$ of the image. At train time, we inserted at most 3 crops per prompt before their corresponding words. We biased our train time data to include images with people, and we also replaced $70\%$ of person crops with a crop of the person's head because we hypothesized face quality would require more higher resolution face training data than clothing or pose quality.

\minihead{synthetic data} We used approximately 3 million synthetic images generated with SDXL and filtered on a minimum PickScore \citep{kirstain2023pick}. Our prompts were composed of a content, e.g. "a man and a dog hiking in the wilderness during the day," concatenated with a style, e.g. "Abstract, geometric, modernist, Cubist influences, bold color blocks, Art Deco elements." To encourage the model to disentangle content and style, we paired each content with many different styles. We used a LLM to extract both contents and styles from DiffusionDB \cite{wang2022diffusiondb}, and we manually prompted a LLM to produce additional contents and styles.

\minihead{realistic data} As SDXL does not produce perfect, high quality, realistic images, we augmented our synthetic data with approximately 2 million high quality, realistic, publicly available images largely containing people. We filtered these images to be SFW, high resolution, non-watermarked, and to contain 0 or 1 people. The images were best effort center cropped to people and captioned with Llava 1.6.

\section{Training Details}
We trained MUMU in two stages on a single 8xH100 GPU node with PyTorch \citep{paszke2019pytorch} FSDP \citep{rajbhandari2020zero,zhao2023pytorch}. 

All images were padded to square resolutions with black pixels. Image crops were always resized up or down to meet their target resolution i.e. we never used less than the target tokens per image. For detected person crops, we replaced the crop of the full person with a crop of their head $70\%$ of the time. Image augmentation of random crops, random flips, Gaussian noise, and grayscaling was used on $20 \%$ of the conditioning images.

\minihead{stage 1} We inserted up to four images per prompt with each individual image using 324 tokens. In each prompt, up to three objects detected in the input image were inserted. $30\%$ of the time we additionally inserted an image of canny edge, depth, or sketch \citep{su2021pixel} of the full input image. We hypothesized this would lead to better ability to combine different image content at test time.

\minihead{stage 2} For each prompt, we inserted a single high resolution face or person crop corresponding to 1,296 tokens to see if more tokens per image improved face quality.

\minihead{hyperparameters} We trained Idefics2 and SDXL with LoRA \cite{hu2021lora} of rank 8 as the models diverged when fully trained. We believe with larger batch sizes, we could fully train all models, and it would result in better model quality. 

\begin{table}[h]
\renewcommand{\arraystretch}{.8}
\centering
\caption{Hyperparameters. The same hyperparameters were used for all training stages.}
\begin{tabular}{lll}
\toprule
\textbf{Model} & \textbf{Training} & \textbf{Learning Rate} \\
\midrule
Idefics2 ViT & Not trained & - \\

Idefics2 ViT to VLM MLP & Rank 8 LoRA & $3e-6$ \\

Idefics2 VLM & Rank 8 LoRA & $3e-6$ \\

Transformer Connector & Fully trained & $1e-5$ \\

SDXL UNet & Rank 8 LoRA & $5e-6$ \\
\bottomrule
\end{tabular}
\label{tab:hyperparameters}
\end{table}

\section{Evaluation}

\begin{figure}[ht!]
    \centering
    \includegraphics[scale=.25]{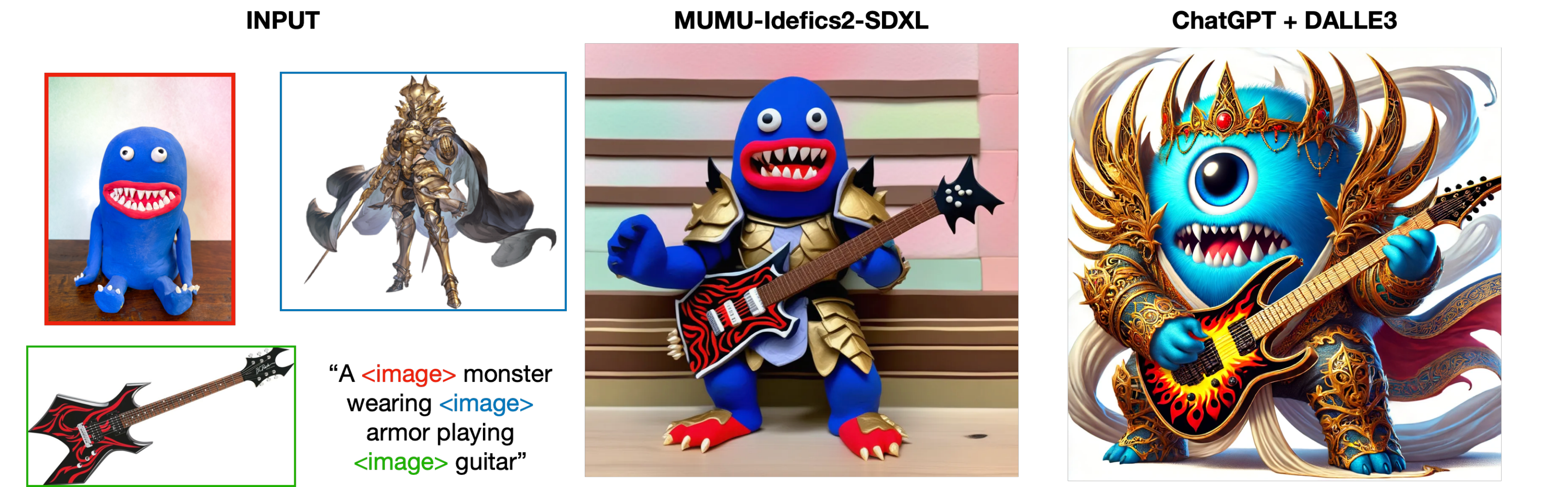}
    \caption{Multimodal prompts with direct inputs of conditioning into the diffusion model (MUMU) allows for much better detail preservation than ChatGPT+DALLE3 which uses images and text to construct a highly detailed text prompt for a text-to-image generator.}
    \label{fig:images_help}
\end{figure}

We now evaluate strengths and weaknesses of MUMU. As a very basic first test, we compare our trained MUMU model to ChatGPT4 + DALLE-3 \citep{betker2023improving}. ChatGPT4 is able to input images, but, as far as we know, only passes text prompts to DALLE-3. In Figure \ref{fig:images_help} we compare a generation from MUMU given the prompt ``a monster wearing armor playing guitar'' and a generation from DALLE-3 when ChatGPT is given the conditioning images and prompted with ``Please generate an image of a monster wearing armor and playing guitar. Attached are images of the monster, armor, and guitar respectively.'' We see that MUMU does a much better job of keeping details of the conditioning images. While MUMU's detail preservation is better than ChatGPT4+DALLE-3, MUMU is also not perfect. One prominent example is that the iconic shape of the guitar in the conditioning image is transformed into a rounded shape in the generated image.

\textbf{Finding 1: Long input contexts are important}
\begin{figure}[ht!]
    \centering
    \includegraphics[scale=.25]{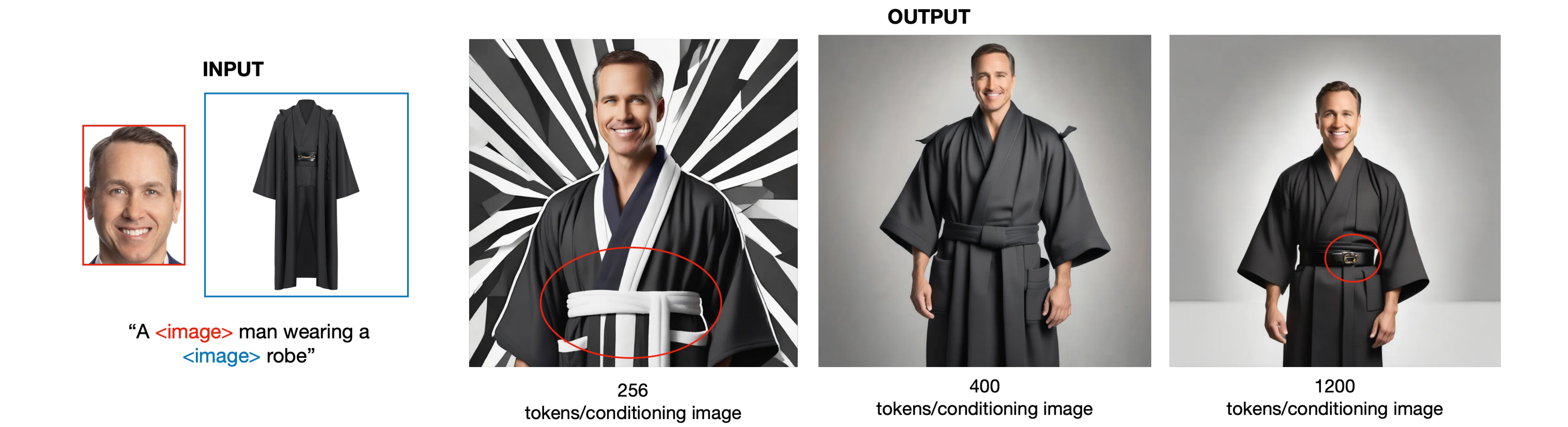}
    \caption{MUMU preserves more detail at higher tokens per image. At lower tokens per image, MUMU captures the gist of `black robe'. At higher tokens per image, details such as the gold inlaid belt are better preserved.}
    \label{fig:num_toks}
\end{figure}

Current VLMs achieve impressive performance with relatively small numbers of tokens per image, e.g. GPT4V uses $85$ + $170$ per $512 \times 512$ tile tokens per image. The baseline version of Idefics2 uses a perceiver to downsample SigLIP tokens to a fixed 64 tokens per image. In early experiments including Idefics2's perceiver, we found that MUMU was good at preserving high level concepts but struggled with details. Detail preservation increases with removing the perceiver and increasing tokens per image up to approximately $1,000$ tokens per image, see Figure \ref{fig:num_toks}.

Using more tokens per image increases training costs, but inference is more reasonable as the encoder is only run once compared to the diffusion model that is iteratively applied at each step.

\textbf{Finding 2: MUMU harmonizes diverse conditioning images}

\begin{figure}[h!]
    \centering
    \includegraphics[scale=.33]{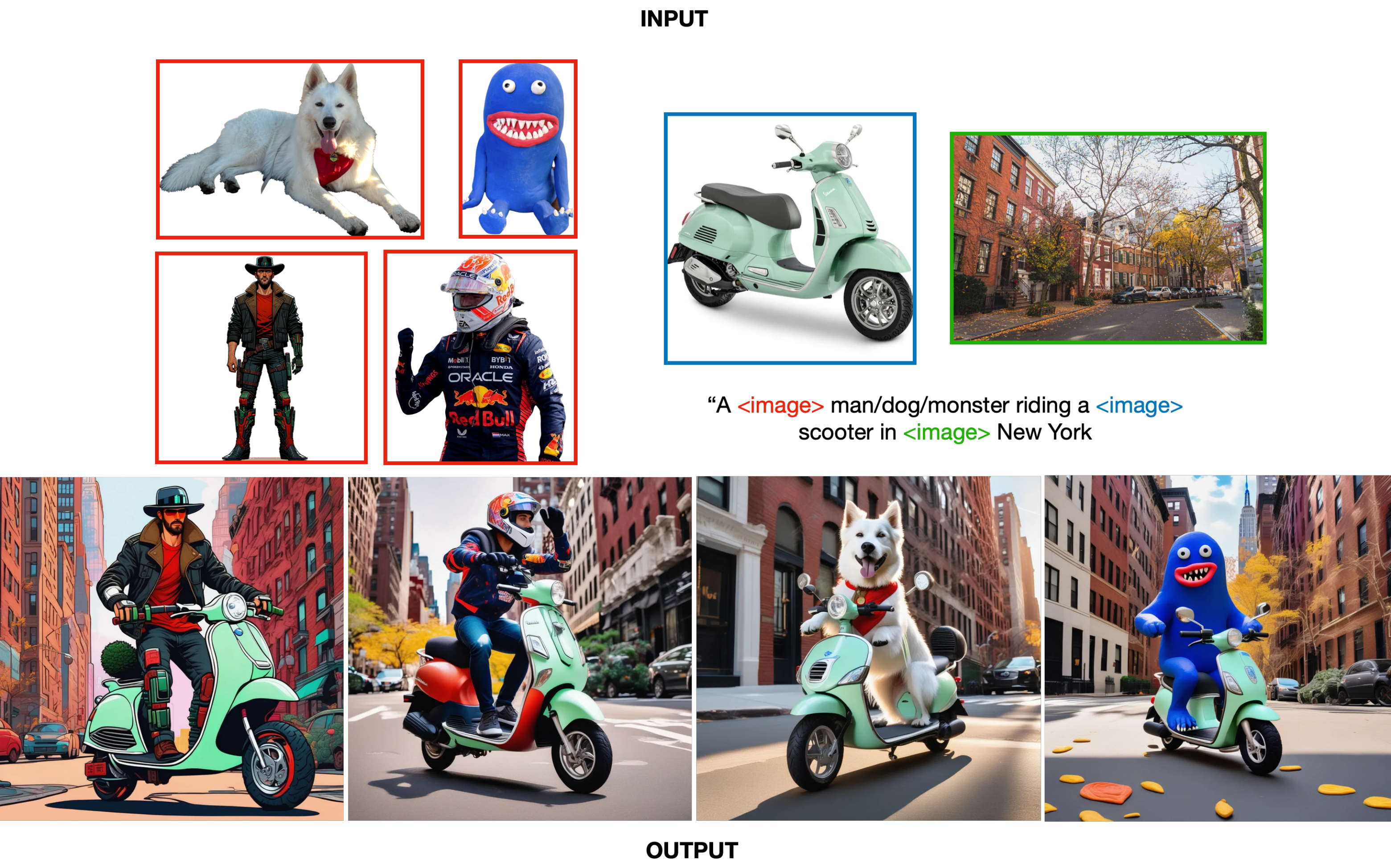}
    \caption{We use a variety of different subjects (dog, race car driver, animated cowboy, blue ceramic monster) in the same prompt ``$\boldsymbol{<}$image$\boldsymbol{>}$ $\boldsymbol{<}$man/dog/monster$\boldsymbol{>}$ riding a $\boldsymbol{<}$image$\boldsymbol{>}$ scooter in $\boldsymbol{<}$image$\boldsymbol{>}$ New York''. MUMU is able to harmonize conditioning image styles and object affordances. We also see that MUMU suffers from common Stable Diffusion artifacts such as concept bleed (e.g. the red on the scooter for the race car driver subject). Blue monster is used with permission from \url{https://www.spacecatceramics.com/}
    }
    \label{fig:consist}
\end{figure}

MUMU is only trained on crops from the same input image. However, as seen in Figure \ref{fig:consist}, the model learns to harmonize conditioning images from different inputs into a coherent generation. E.g. an input of a realistic person and a cartoon will output the same person in the cartoon style. Additionally, MUMU harmonizes object affordances. An input of a standing subject\footnote{Blue monster is used with permission from \url{https://www.spacecatceramics.com/}.} and a scooter will output the subject riding the scooter. 

\textbf{Finding 3: MUMU can perform some style transfer}

We also investigated pure style transfer rather than simple image harmonization. In style transfer, the reference object is meant to serve as a global style reference rather than be placed in the generation. This task is quite far outside of the model's training distribution.

\begin{figure}[h!]
    \centering
    \includegraphics[scale=.3]{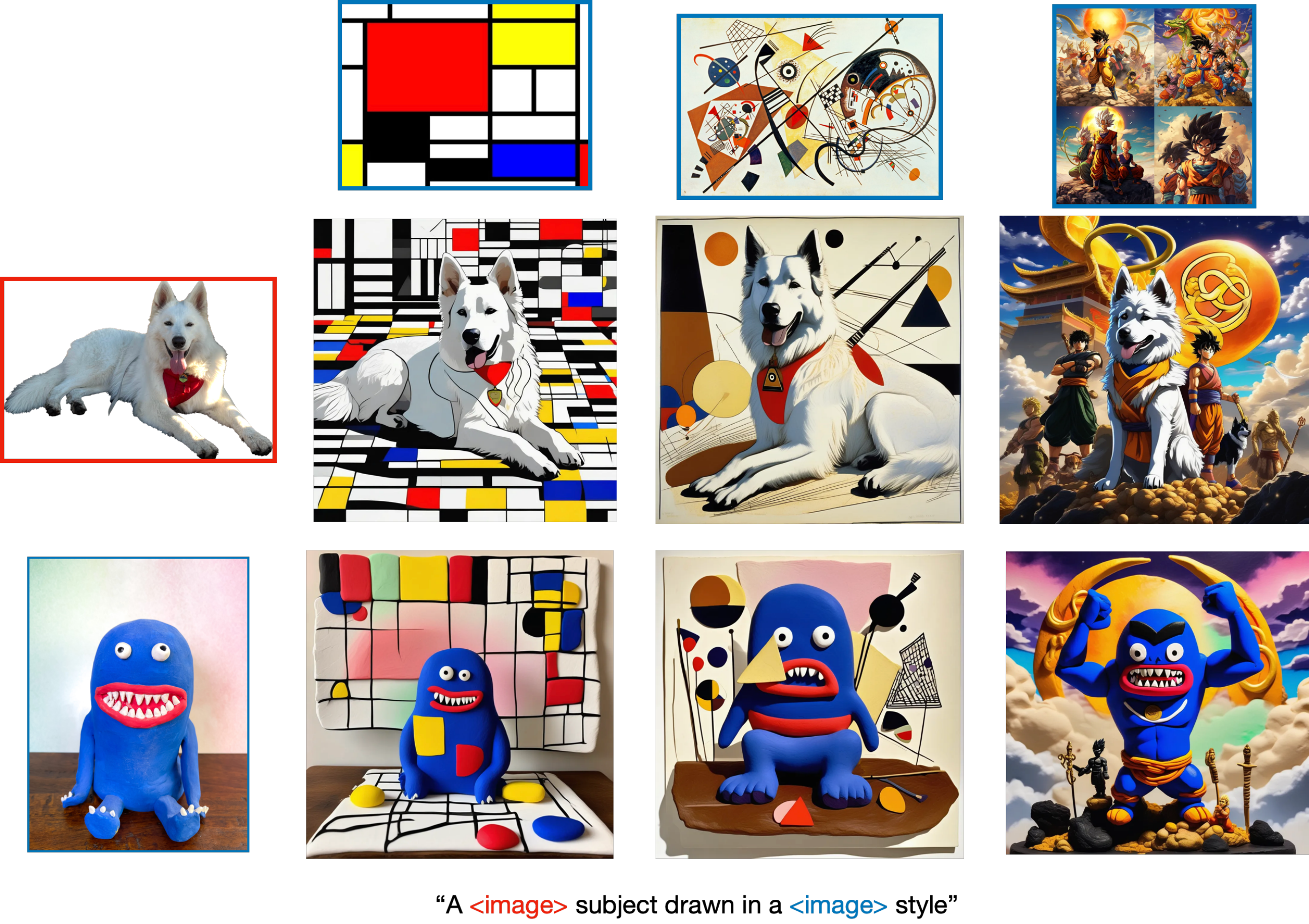}
    \caption{MUMU harmonization means the model is able to do some amount of style transfer.}
    \label{fig:styletransfer}
\end{figure}

\begin{figure}[h!]
    \centering
    \includegraphics[scale=.3]{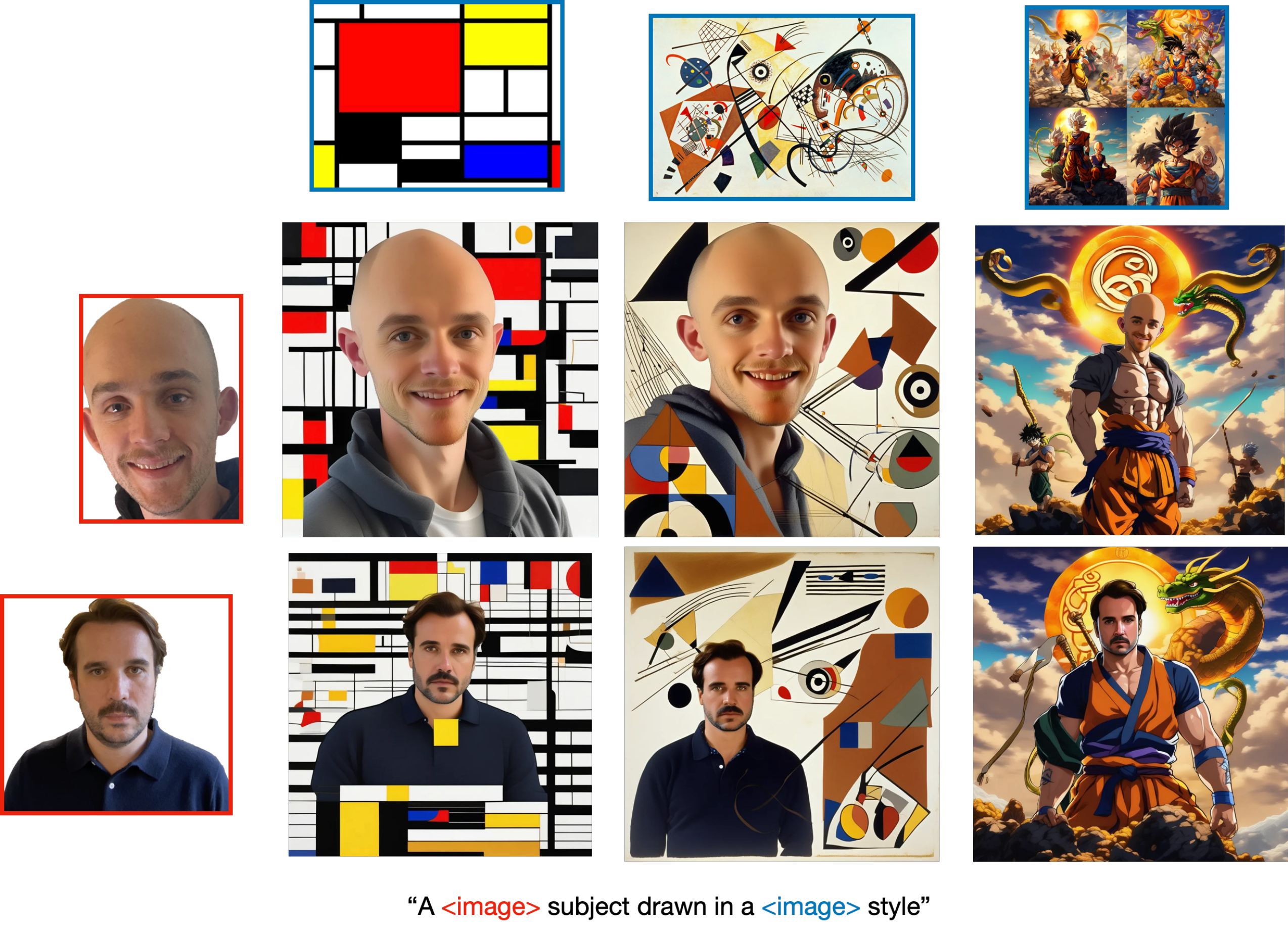}
    \caption{Style transfer for human subjects combined with very abstract styles is noticeably worse in our model. This is likely due to our special emphasis on faces in training.}
    \label{fig:stylefail}
\end{figure}

We found that MUMU can somewhat perform style transfer though it is not perfect. Figure \ref{fig:styletransfer} shows more successful examples of style transfer while Figure \ref{fig:stylefail} shows less successful ones. 

In general, we found that a major failure point of MUMU was an inability to translate human faces into abstract styles. We hypothesize this is because of oversampling human head/face conditioning at train time.
 
\textbf{Finding 4: Community SDXL fine-tunes can be model merged with MUMU}

\begin{figure}[h!]
  \centering
  \includegraphics[scale=.3]{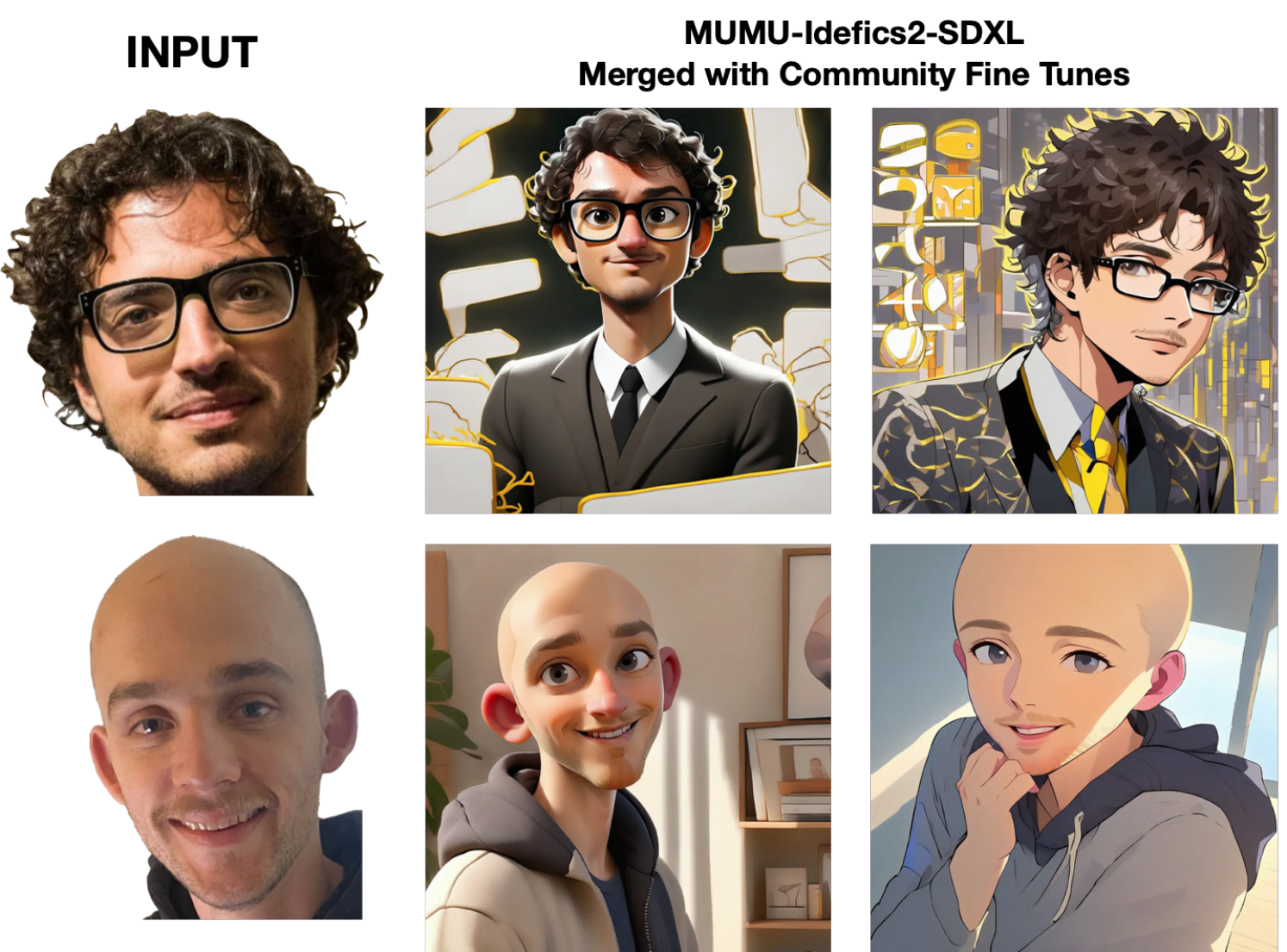}
  \caption{MUMU is compatible with existing SDXL community style fine-tunes.}
  \label{fig:lora}
\end{figure}

SD1.5 and SDXL have vibrant communities that create base model fine-tunes. We explored whether MUMU's LoRAs can be used with style specific fine-tunes without re-training. To do this, we merged the MUMU LoRA with community SDXL fine-tunes\footnote{\url{https://huggingface.co/cagliostrolab/animagine-xl-3.1} \url{https://huggingface.co/stablediffusionapi/samaritan-3d-cartoon}} to generate the prompted images in the fine-tune style. We see in Figure \ref{fig:lora} that the MUMU LoRA appears to compose with the stylistic fine-tunes though more work is needed to see whether any quality degradation occurs compared to the baseline.

\textbf{Finding 5: More work is needed for detail consistency}

Prior examples show that even at large numbers of image tokens, MUMU-Idefics2-SDXL does not achieve perfect consistency in preserving details. 

We now specifically consider the problem of character consistency. Figure \ref{fig:consist} shows several portraits generated by MUMU from conditioning images of the authors compared to InstantID \citep{wang2024instantid}.\footnote{The InstantID examples are done via the official Huggingface Demo \url{https://huggingface.co/spaces/InstantX/InstantID} with all default parameters, no `style' setting, an a prompt of `a portrait of a person.' More hyper parameter tuning could increase aesthetic quality of the InstantID output.} InstantID is an adapter and ControlNet based SDXL fine-tune specialized for face generation. InstantID uses face embeddings, face key points, and text all as inputs. We see that InstantID does a better job at preserving facial features than MUMU, but a worse job at preserving other details (e.g. hair). We believe that further improving the vision model in a MUMU-like architecture an is a key area for improvement of facial consistency.
 
\begin{figure}[h!]
    \centering
    \includegraphics[scale=.3]{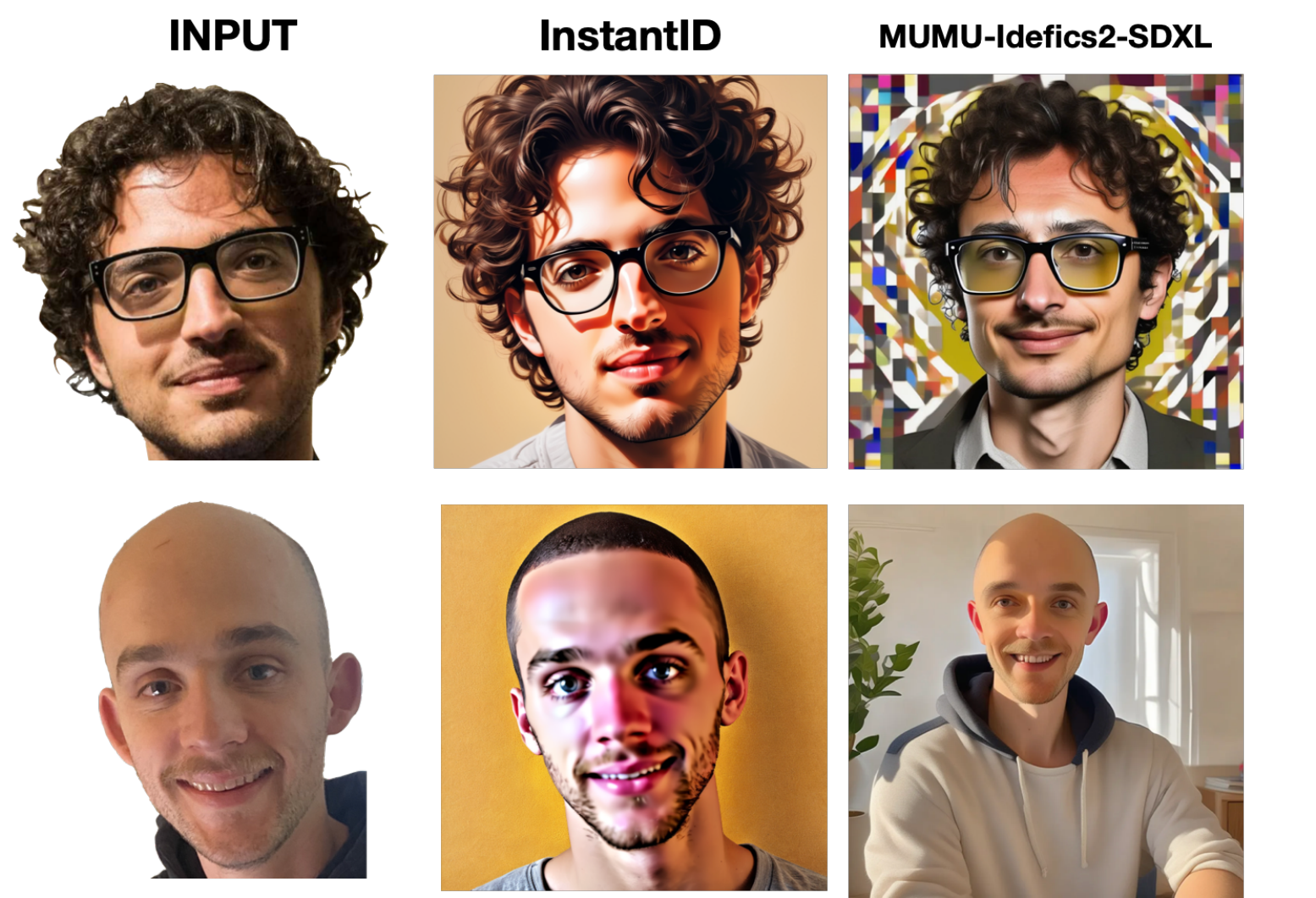}
    \caption{
    MUMU, using $\sim 1000$ tokens per image with the prompt ``a <image> portrait'', compared to InstantID. MUMU misses important details and has worse ability to capture the face detail. However, MUMU can better capture non-face detail like hair and glasses.
    }
    \label{fig:consist}
\end{figure}

\textbf{Cherry Picking}

We present MUMU with examples rather than quantitative evaluations claiming `SoTA' on specific tasks. We want to be transparent about MUMU's failure cases and how much cherry picking was involved in choosing examples. MUMU has generally coherent outputs with typical Stable Diffusion artifacts such as extra objects, extra limbs, missing limbs, and concept bleeding. We chose all figures from $1-5$ generations with a guidance scale of $6$ and no negative prompt. We did no finetuning on specific or out of distribution subjects or prompts.

\begin{figure}[h!]
    \centering
    \includegraphics[scale=.3]{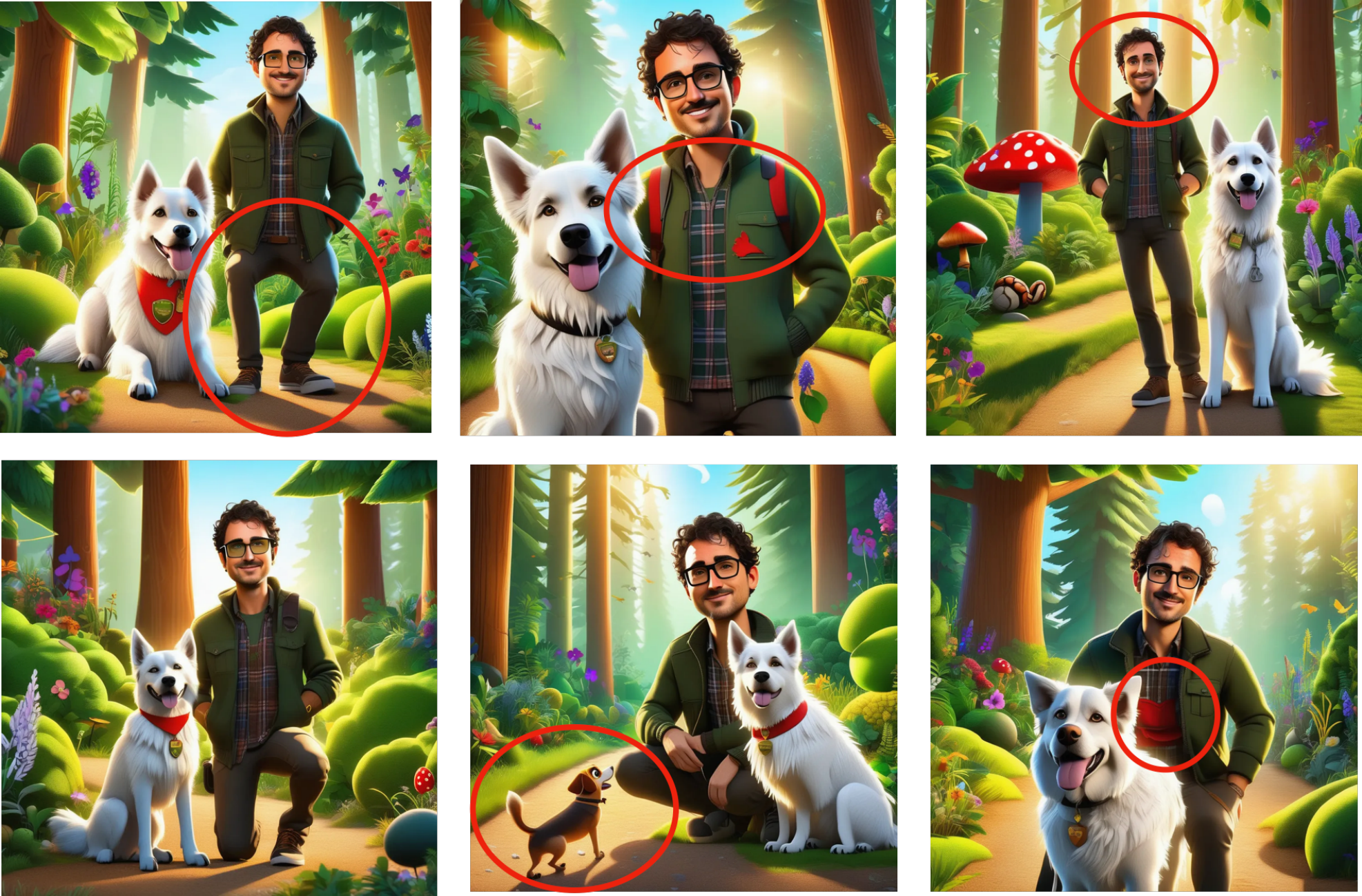}
    \caption{
    We sample the prompt from Figure \ref{fig:first_example}, “a $\boldsymbol{<}$picture of a man$\boldsymbol{>}$ man and his $\boldsymbol{<}$picture of a dog$\boldsymbol{>}$ dog in an $\boldsymbol{<}$picture of a cartoon$\boldsymbol{>}$ animated style,” $6$ times. MUMU has generally coherent outputs with typical Stable Diffusion artifacts. 
    }

    \label{fig:cherry}
\end{figure}

\textbf{Summary}

MUMU-Idefics2-SDXL is not "production ready," but its ability to generalize outside the training task shows the promise of multimodal models as general controllers for image generation. We believe many failure points presented here can be remedied simply with scale: increasing the data set, using more detailed captions, training for longer with bigger batch sizes, full training instead of using LoRA, and unfreezing the vision tokenizer. We leave scaling to the GPU rich.

\section{Some Open Questions Beyond Scaling}
There are many directions beyond `make it bigger' that we did not get a chance to explore. We outline some of them here in hopes of stimulating ideas for further research.

\textbf{Question 1: Architecture choices}

\minihead{lora training} We hypothesize the largest increase in model quality would come from fully training all models without LoRAs. We found that at our batch sizes and learning rates fully training the models let to divergences, so we leave this scaling to further work.

\minihead{image tokenization} It is not clear that SigLIP's text-image discrimination pre-training objective and Idefics2's VQA finetuning objective are optimal for preserving fine image details for generation. We believe alternative image tokenization methods or directly training the image encoder could improve model quality.

\minihead{token based decoder} We use a diffusion decoder because SDXL is the highest quality open source image model. However, MUMU is decoder agnostic and could be retrained with an autoregressive or other token-based image generation backbone \citep{ramesh2021zero,yu2023language,chang2022maskgit,yu2023scaling, yu2022scaling,team2024chameleon,team2023gemini,aiello2023jointly,openai2024gpt4o}. Unfortunately, most of these backbones are closed source, and open source token based models \citep{patil2024amused,esser2021taming} generate lower quality images than SDXL.

\textbf{Question 2: Data}

We view the dataset image composition and annotations as perhaps the most important direction for MUMU. Our dataset is more than half synthetic data with relatively short prompts, and our small VLM generates relatively short and coarse captions for our non-synthetic data. Longer quality captions improve text-to-image models \cite{betker2023improving,esser2024scaling,chen2024pixartsigma}, and so we believe that longer multimodal captions with more text, more objects, and other information that is readily extractable and tokenizable (for example, spatial location of each object) could also improve MUMU. 

In addition, we include object crops directly in the multimodal prompt. At test time, we typically use segmented objects with the background removed because leaving in the background causes the model to use background details in the generation. Segmenting objects rather than simply cropping them at training time (e.g. by using \citep{kirillov2023segment}) is an interesting extension.

\textbf{Question 3: Evaluation}
As there are no widely agreed upon metrics for multimodal prompted image models, we used qualitative evaluations in this paper. We constructed a multimodal prompt CLIP score measuring multimodal prompt-image alignment, but it was not a useful guide for model quality.

\begin{figure}[h!]
\centering
  \includegraphics[scale=.17]{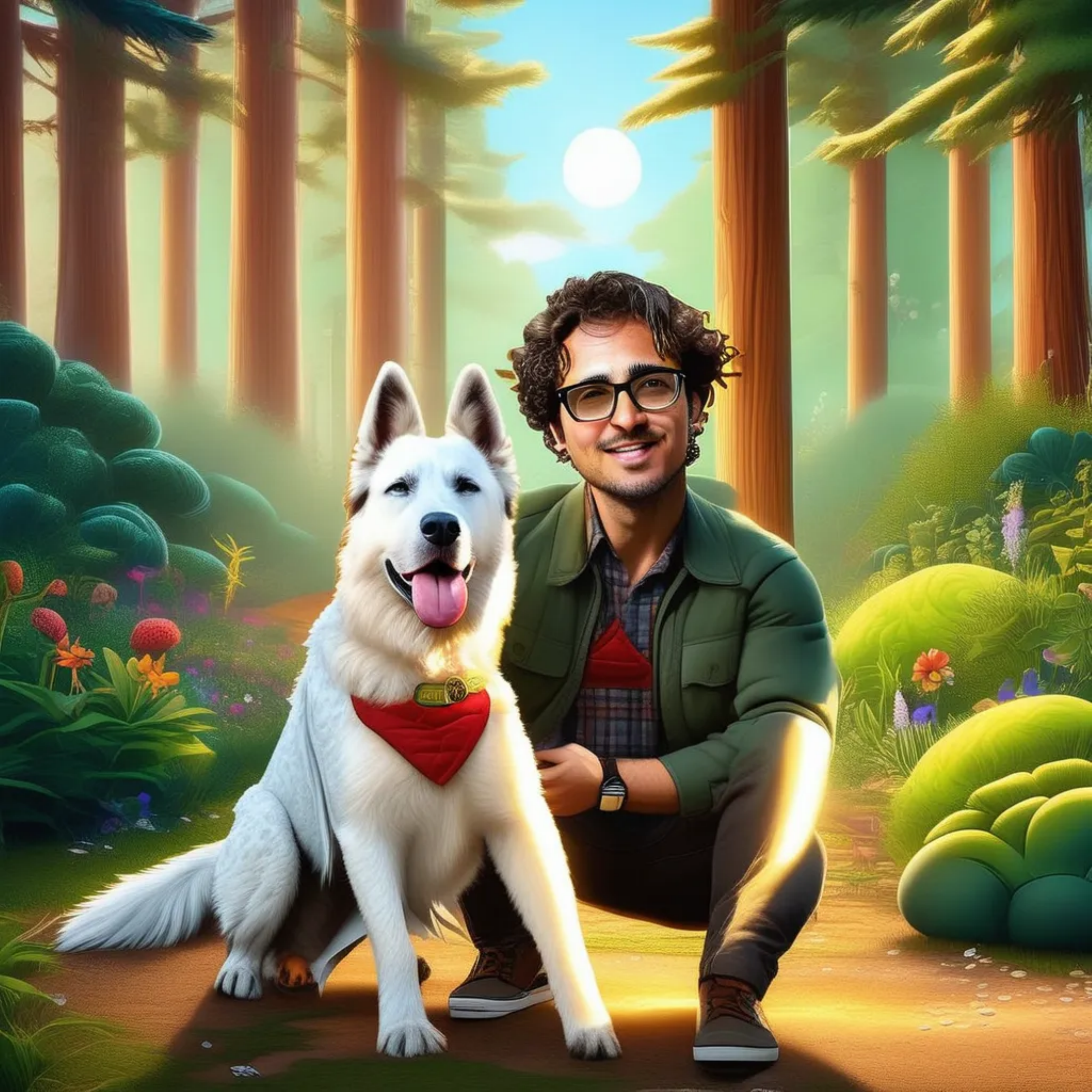}
  \caption{
  Models which did well on multimodal prompt-image CLIP score and face embedding similarity did not harmonize diverse inputs as well. This example uses the same prompt, “a $\boldsymbol{<}$picture of a man$\boldsymbol{>}$ man and his $\boldsymbol{<}$picture of a dog$\boldsymbol{>}$ dog in an $\boldsymbol{<}$picture of a cartoon$\boldsymbol{>}$ animated style,” from Figure \ref{fig:first_example}.
  }
  \label{fig:no_harmonization}
\end{figure}

Our multimodal prompt-image CLIP score computed average CLIP similarity between objects in the input prompt and corresponding objects in the output image. However, it did not help us make good experimental decisions. For example, removing data augmentation created models with strong multimodal CLIP scores, but they had a tendency to copy-paste inputs directly into the generation and were not as good at harmonizing diverse inputs, see Figure \ref{fig:no_harmonization} for an example output.

We also measured face consistency with the face embedding from \citep{wang2024instantid}, but models with high scores on this metric similarly were not as good at harmonizing input images.

Most likely good harmonization comes at some amount of detail loss. Understanding how to measure and balance these two objectives is an important avenue for future work.

\section{Conclusion}
We have demonstrated a method for bootstrapping multimodal prompts from a text, image dataset, and we have trained a multimodal prompted image generation model with off-the-shelf encoders and decoders.

We view multimodal inputs as an important step for any application of generative AI, and we have discussed many interesting directions for future research. Multimodal inputs unlock the possibility for users to not simply guess at what text will generate. For example, a user could hand draw a single image of a character, and then generate the character in various poses and environments. Ultimately we hope that our work contributes to the important task of making a `creative copilot'.

\bibliographystyle{unsrt} 
\bibliography{main.bbl}

\end{document}